\documentclass[10pt,twocolumn,letterpaper]{article}

\usepackage{cvpr}
\usepackage{times}
\usepackage{epsfig}
\usepackage{graphicx}
\usepackage{amsmath}
\usepackage{amssymb}
\usepackage[ruled,vlined]{algorithm2e}
\usepackage{multirow}
\usepackage{bigstrut}
\usepackage[font=footnotesize]{caption}
\usepackage{subfig}
\usepackage{caption}
\usepackage{colortbl}
\usepackage[pagebackref=true,breaklinks=true,letterpaper=true,colorlinks,bookmarks=false]{hyperref}

\cvprfinalcopy % *** Uncomment this line for the final submission

\def\cvprPaperID{3278} % *** Enter the CVPR Paper ID here

\ifcvprfinal\fi
\begin{document}

%%%%%%%%% TITLE
\title{Kernel Cross-View Collaborative Representation based Classification for Person Re-Identification}

\author{Raphael Prates and William Robson Schwartz\\
Universidade Federal de Minas Gerais, Brazil\\
6627, Av. Pres. Ant\^onio Carlos - Pampulha, Belo Horizonte - MG, 31270-901\\
{\tt\small prates@dcc.ufmg.br, william@dcc.ufmg.br}
}

\maketitle
%\thispagestyle{empty}

%%%%%%%%% ABSTRACT
\begin{abstract}
Person re-identification aims at the maintenance of a global identity as a person moves among non-overlapping surveillance cameras. It is a hard task due to different illumination conditions, viewpoints and the small number of annotated individuals from each pair of cameras (small-sample-size problem). Collaborative Representation based Classification (CRC) has been employed successfully to address the small-sample-size problem in computer vision. However, the original CRC formulation is not well-suited for person re-identification since it does not consider that probe and gallery samples are from different cameras. Furthermore, it is a linear model, while appearance changes caused by different camera conditions indicate a strong nonlinear transition between cameras. To overcome such limitations, we propose the \textit{Kernel Cross-View Collaborative Representation based Classification} (\textit{Kernel X-CRC}) that represents probe and gallery images by balancing representativeness and similarity nonlinearly. It assumes that a probe and its corresponding gallery image are represented with similar coding vectors using individuals from the training set. Experimental results demonstrate that our assumption is true when using a high-dimensional feature vector and becomes more compelling when dealing with a low-dimensional and discriminative representation computed using a common subspace learning method. We achieve state-of-the-art for~\emph{rank-1} matching rates in two person re-identification datasets (PRID450S and GRID) and the second best results on VIPeR and CUHK01 datasets.
\end{abstract}

%%%%%%%%% BODY TEXT
\section{Introduction}

% Person Re-Identification
%Re-identifying people across non-overlapping cameras, known in the literature as the 

Person re-identification (Re-ID) plays a key role in security management applications and has received increasing attention in the past years~\cite{paper38}. Its goal is to identify a person (probe sample), captured by one or more cameras, using a gallery of  already known candidates captured from a different camera. Most of the works consider the single-shot and two surveillance cameras scenario, where a single subject image is available for each camera. The restricted number of samples and cameras makes the problem more challenging due the small-sample-size problem~\cite{paper52}. Despite the efforts from the computer vision community, Re-ID remains an unsolved problem due to appearance changes caused by pose, occlusion, illumination and camera transition.  %Figure~\ref{fig:fig1} illustrates some of these problems.

%\begin{figure}[!t]
%\centering
%\includegraphics[width=3.5in]{figures/fig1}
%\caption{Images of the same individual (columns) captured by cameras A (first row) and B (last row), in VIPeR dataset. Notice that the same person looks dissimilar when captured by cameras with distinct conditions. We can attribute these appearance changes to different pose (frontal and back), camera viewpoints, illumination conditions (sunny and shadow regions), self-occlusion and cluttered background.}
%\label{fig:fig1}
%\end{figure}

Due to the lack of available samples or to the high cost for collecting and annotating a large number of images for each subject in the gallery, the small-sample-size problem also appears on the face identification task. To handle this problem, researchers have focused on Sparse Representation based Classification (SRC)~\cite{paper53} and Collaborative Representation based Classification (CRC)~\cite{paper52} by representing each probe image $\textbf{y}$ using all the images in the gallery $\textbf{X}$ with a proper regularization term as
%In face recognition, specifically on the face identification task, the goal is to recognize a probe sample among a gallery of known individuals. However, due to lack of available samples or the high cost for collecting and annotating a large number of images for each individual in the gallery, the researchers also face the small-sample-size problem. Recently, Sparse Representation based Classification (SRC)~\cite{paper53} and Collaborative Representation based Classification (CRC)~\cite{paper52} addressed this problem representing each probe image  $\textbf{y}$ using all the images in the gallery $\textbf{X}$ with a proper regularization term as
\begin{equation}
\min_\alpha \parallel \textbf{y} - \textbf{X} \pmb{\alpha}\parallel_2^2 +\lambda\parallel\pmb{\alpha}\parallel_{p},
\label{eq:eq01}
\end{equation}
where $\lambda$ is a scalar and $\pmb{\alpha}$ is the sparse ($p=1$) or the collaborative ($p=2$) coding vector~\cite{paper52,paper53}. These methods assign the probe image to the class that results in the smallest reconstruction error. With this approach, researchers have achieved high performance in applications such as face recognition~\cite{paper52,paper54,paper59,paper63,paper53}, hyperspectral image classification~\cite{paper62} and multimodal biometrics~\cite{paper64}. 

\begin{figure*}
  \begin{minipage}{.5\textwidth}
  \centering
    \includegraphics[width=.93\textwidth]{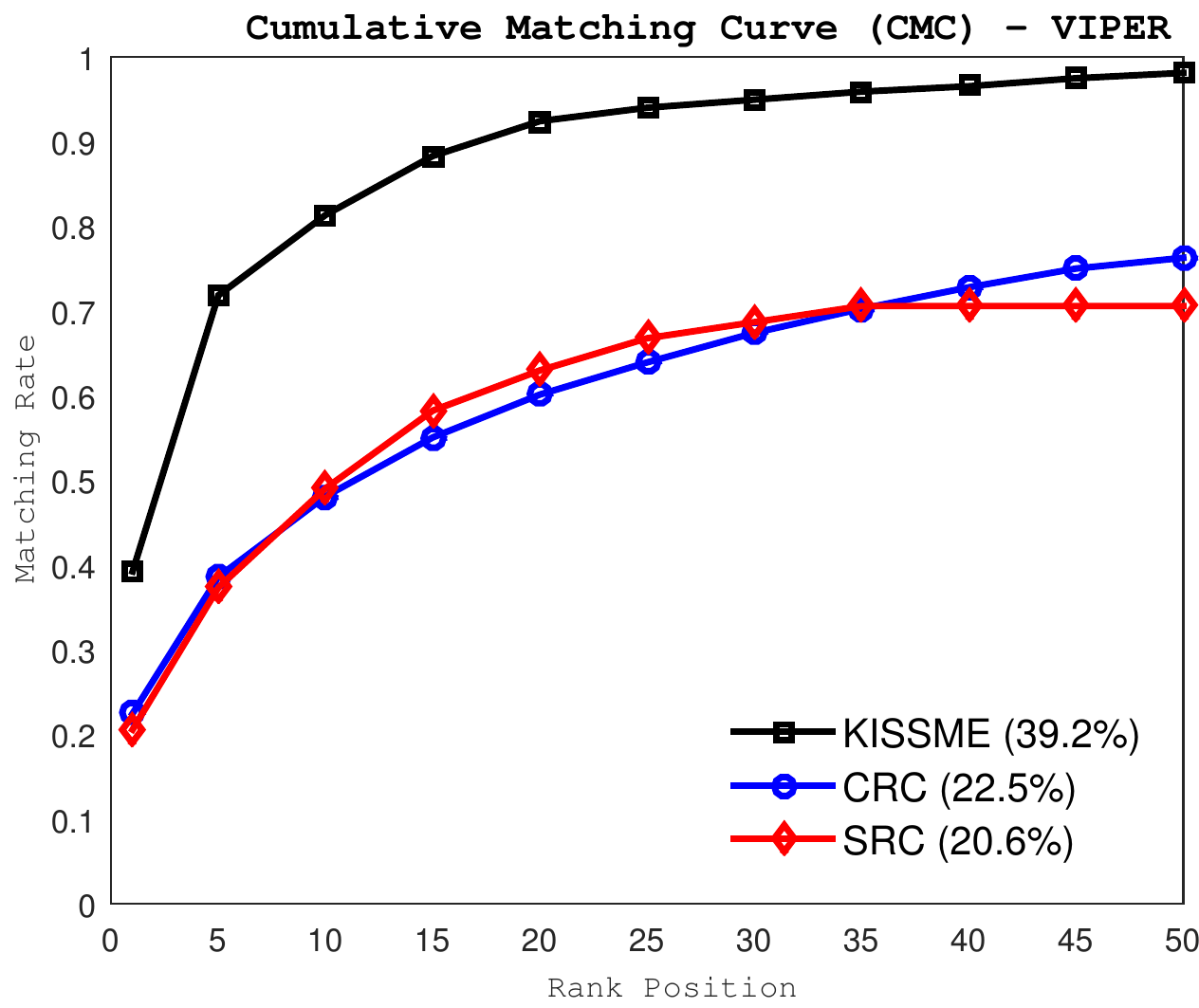}
   \end{minipage}%
  \begin{minipage}{.5\textwidth}
  \centering
    \includegraphics[width=.93\textwidth]{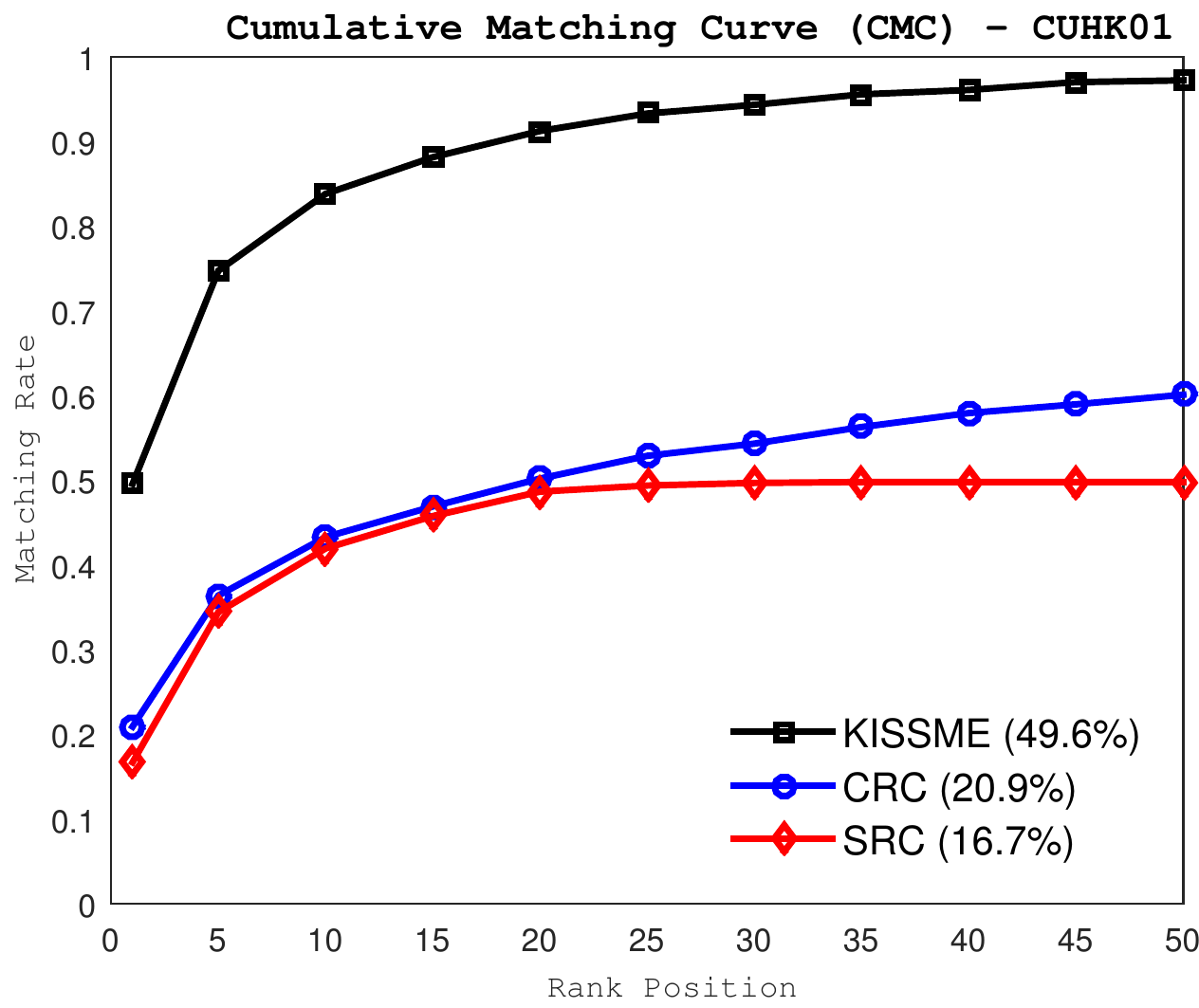}
   
  \end{minipage}
  \caption{Comparison between Sparse Representation based Classification (SRC), Collaborative Representation based Classification (CRC) and KISSME~\cite{paper71} on two widely used person Re-ID datasets (VIPeR~\cite{paper68} and CUHK01~\cite{paper67}). The values in parenthesis indicate the~\emph{rank-1} matching rate.}
  \label{fig:baselines}
\end{figure*}

Despite the accurate results in different computer vision problems, experimental results demonstrate that SRC and CRC classifiers achieve weak matching performance on the person re-identification problem when compared to a baseline obtained by the well-known KISSME~\cite{paper71} approach, as shown in Figure~\ref{fig:baselines} for two widely used Re-ID datasets. While KISSME uses the available image pairs in the training set to learn a discriminative cross-view metric distance, SRC and CRC compare the probe and gallery images directly based on the reconstruction error disregarding the camera transition, which is one of the main challenges of Re-ID. Therefore, the following questions should be addressed.  How to include discriminative cross-view information in a collaborative representation framework? How to indirectly compare probe and gallery images?

An attempt to adapt the collaborative representation to the person Re-ID problem would be to compute the collaborative representation coefficients (coding vectors) $\pmb{\alpha}_x$ and $\pmb{\alpha}_y$ by solving the camera-specific optimization problems 
\begin{equation}
\begin{split}
\min_{\pmb{\alpha}_y} \parallel \textbf{y} - \textbf{D}_y \pmb{\alpha}_y\parallel_2^2 +\lambda\parallel\pmb{\alpha}_y\parallel_{2}^2 \text{ and} \\
\min_{\pmb{\alpha}_x} \parallel \textbf{x} - \textbf{D}_x \pmb{\alpha}_x\parallel_2^2 +\lambda\parallel\pmb{\alpha}_x\parallel_{2}^2,
\end{split}
\label{eq:eq02}
\end{equation}
\noindent where each linear regularized model represents the probe sample, $\textbf{y}$,  or the gallery image,  $\textbf{x}$, using the respective images in the training set ($\textbf{D}_x$ or $\textbf{D}_y$). The feature descriptors extracted from a subject in the training set captured by probe and gallery cameras are assigned to each column of $\textbf{D}_y$ and $\textbf{D}_x$, respectively. Therefore, similar coding coefficients would be expected when $\textbf{x}$ and $\textbf{y}$ correspond to a same individuals acquired from different cameras. The matching between probe and gallery images occurs indirectly using the coding vectors ($\pmb{\alpha}_x$ and $\pmb{\alpha}_y$). Note that $\textbf{D}_x$ and $\textbf{D}_y$ can be any representation of the training images able to balance discriminative power and robustness to camera transition.

%We named aforementioned camera-specific approach \emph{Cross-View Collaborative Representation based Classification (C\textsuperscript{2}RC)}. \todo{ficou estranho, voce fala que foi proposto C\textsuperscript{2}RC, nao ha referencia e depois nao fala mais sobre isso no paper e foca no kernel xcrc. Talvez o melhor seria descrever essa abordagem como uma possibilidade, mas sem dar nome e jah comecaria falar das desvantagens dela e passaria ao kernel xcrc}. However,

The limitation of Equation~\ref{eq:eq02}, however, is that it only considers the representativeness in the camera-specific optimization problems and disregards that elements $\pmb{\alpha}_x^i$ and $\pmb{\alpha}_y^i$ contain information about the \textit{i-th} subject from training set, which we expect to be similar when $\textbf{x}$ and $\textbf{y}$ represents the same individual captured from different camera-views. Therefore, a better model should consider the balance between the representativeness and similarity when computing $\pmb{\alpha}_x$ and $\pmb{\alpha}_y$ in a unique multi-task framework. In addition, previous works demonstrated improved performance when handling the nonlinear behavior of data using kernel functions instead of the linear modeling used in Equation~\ref{eq:eq02}~\cite{paper08,paper79,paper80}.

In this work, we propose a novel method to address the person Re-ID problem using a supervised CRC framework, the~\emph{Kernel Cross-View Collaborative Representation based Classification  (Kernel X-CRC)}, inspired on Collaborative Representation based Classification (CRC) in the sense that it has a analytical solution that represents each pair probe $\textbf{y}$ and gallery $\textbf{x}$ images collaboratively using its camera-view specific training samples $\textbf{D}_y$ and $\textbf{D}_x$, respectively. Differently from Equation~\ref{eq:eq02}, the coding vectors $\pmb{\alpha}_x$ and $\pmb{\alpha}_y$ are computed in a unique multi-task learning framework.  

A multi-task learning framework provides evidences that related tasks transfer knowledge when learned simultaneously, improving the generalization performance~\cite{paper54,paper50,paper60,paper61}. Considering the computation of the collaborative representation coefficients in each camera-view as a different task, we estimate the coding vectors $\pmb{\alpha}_x$ and $\pmb{\alpha}_y$ simultaneously with a similarity term to balance the intra-camera representativeness and the inter-camera discriminative power. Furthermore, we learn $\pmb{\alpha}_x$ and $\pmb{\alpha}_y$ in a nonlinear feature space to be able to handle the strong nonlinearity present on the person re-identification data~\cite{paper08,paper79,paper80}. Therefore, the \emph{Kernel X-CRC} computes a multiple regularized linear model with proper regularization of parameters for each pair of probe and gallery images, as
\small
\begin{equation}
\begin{split}
\min_{\pmb{\alpha}_x,\pmb{\alpha}_y}\parallel\phi(\textbf{y}) - \pmb{\Phi}_y\pmb{\alpha}_y\parallel_2^2  +\parallel\phi(\textbf{x}) -\pmb{\Phi}_x \pmb{\alpha}_x\parallel_2^2 \\
+\lambda\parallel\pmb{\alpha}_y\parallel_2^2 + \lambda\parallel\pmb{\alpha}_x\parallel_2^2 + \parallel\pmb{\alpha}_y -\pmb{\alpha}_x\parallel_2^2,
\end{split}
\label{eq:eq02b}
\end{equation}
\normalsize
\noindent where $\phi(.)$ is a nonlinear function and, $\pmb{\Phi}_x$ and $\pmb{\Phi}_y$ are resulting nonlinear mapping of $\textbf{D}_x$ and $\textbf{D}_y$, respectively. 

Despite its simplicity, the~\emph{Kernel X-CRC} successfully balances the representativeness and similarity to obtain nonlinear and discriminative coding vectors for each pair of probe and gallery images. Then, by matching the computed coding vectors using a simple cosine distance, we obtain improved results when compared to state-of-the-art methods. According to experimental results, the proposed~\emph{Kernel X-CRC} outperforms state-of-the-art single-shot based methods in the smaller datasets evaluated (PRID450S and GRID), where the small-sample-size problem is more critical, and is also successful in the larger datasets, holding the second best performance in VIPeR and CUHK01 datasets.

%More specifically, our experiments demonstrate that when representing training samples $\textbf{D}_y$ and $\textbf{D}_x$ in a low-dimensional common subspace \done{computed using the Cross-View Quadratic Discriminant Analysis method (XQDA~\cite{paper45})}, the proposed~\emph{Kernel X-CRC} outperforms state-of-the-art single-shot based methods in the smaller datasets evaluated (PRID450S and GRID), where the small-sample-size problem is more critical. Furthermore,~\emph{Kernel X-CRC} is also successful in the larger datasets, holding the second best performance in VIPeR and CUHK01 datasets.\done{It is important to highlight that we used XQDA~\cite{paper45} due the online available code and the outperforming results. However, the proposed~\emph{Kernel X-CRC} is a general method that can be employed with the original feature representation or any other common subspace learning method.}

\vspace{1mm}
\noindent\textbf{Contributions.} The main contributions of this work are the following. We propose a novel approach to the Re-ID problem using a~\emph{Kernel Cross-View Collaborative Representation based Classification  (Kernel X-CRC)} that embodies cross-view discriminative information and models the nonlinear behavior of person re-identification data. Furthermore, we present an efficient analytical solutions to~\emph{Kernel X-CRC} with outperforming results when compared with more complex state-of-the-art approaches at four challenging datasets (VIPeR, PRID450S, CUHK01 and GRID). 

To the best of our knowledge, this is the first work addressing the person re-identification problem as a multi-task collaborative representation problem. Furthermore, it is important to emphasize that, even though employed to Re-ID problem, the proposed approach provides a general framework that does not consider any extra information at testing stage and could also be employed to other computer vision problems, such as face recognition, hyperspectral image classification and multi-modal biometrics.
\section{Related Work}

%This section reviews the main works focused on person re-identification. For a more detailed discussion, please refer to~\cite{paper38}. Furthermore, we present a brief revision of related works that use joint sparse or collaborative representations to combine complementary modalities or feature descriptors. 

Different feature descriptors have been proposed for the Re-ID problem by exploiting feature representation and body parts from where they are extracted and matched~\cite{paper40,paper48,paper42,paper43,paper46,paper18,paper41,paper44,paper45,paper07,paper72,paper81,paper83,paper90}. Regarding the body locations, Farenzena et~\textit{al.}~\cite{paper48} use human symmetry to determine discriminative body locations, while body parts are detected using Pictorial Structures in~\cite{paper42}. Similarly, Cai and Pietikainen~\cite{paper43} extract descriptors from fixed log-polar grids and Liao et~\textit{al.}~\cite{paper45} construct a stable representation using local patches obtained in multiple scales. A different approach constraints the inter-camera matching computing patch saliency information~\cite{paper46} or capturing spatial distribution of patches between cameras~\cite{paper83}. 

With respect to the feature representation, some works capture information using Fisher Vectors~\cite{paper40}, histograms of semantic color names~\cite{paper44}, represent local patches with hierarchical Gaussian distribution~\cite{paper81} and design biologically inspired features and covariance descriptors~\cite{paper41}. To handle the camera transition, Chen et~\textit{al.}~\cite{paper90} propose to align feature distribution across disjoint using the Mirror Representation~\cite{paper90}, while the feature importance can be computed learning a fixed statistical model~\cite{paper07} or adaptively from subsets of similar individuals and random forests~\cite{paper70}. More recently, Wu~et~\textit{al.}~\cite{paper72} combine hand-crafted and deep learning-based feature descriptors to obtain a discriminative deep feature representation. In this work, we assume that it is not possible to properly handle the camera transition by directly matching feature descriptors captured by different cameras. Therefore, we use a supervised nonlinear framework to compute coding vectors that capture cross-view discriminative information to perform an indirect matching between probe and gallery images.

In supervised learning-based approaches, feature descriptors are combined with discriminative models learned using labeled images from camera pairs to obtain higher matching performance. For instance, distance metric learning-based approaches use the pairwise constraint to learn a distance function that is smaller between pairs of the same person and larger otherwise~\cite{paper71,paper73,paper74,paper84,paper85,paper86,paper87,paper93}. As example, WARCA~\cite{paper84} learns a Mahalanobis distance in a low-dimensional subspace computed using orthonormal regularizer, while NLML~\cite{paper85} learns multiple sets of nonlinear transformations using feed-forward neural network and large margin optimization. Differently, Cheng et~\textit{al.}~\cite{paper89} use a triplet loss function that keeps closer instances of the same person in features space learned using multi-channel CNN. Zhang et~\textit{al.}~\cite{paper91} present a distinct approach that relates the model parameters and the feature space using semi-coupled dictionary learning to obtain a model specific for each individual representation. Similarly, in this work we compute models specific for each pair of probe and gallery images. However, we do not assume that it is possible to learn an effective mapping function between feature space and parameters space using a reduced number of training samples.

Subspace learning methods have been widely employed by supervised Re-ID approaches~\cite{paper04,paper05,paper08,paper11,paper45,paper79,paper80,paper82}. An et~\textit{al.}~\cite{paper05} use Canonical Correlation Analysis (CCA) to learn projections to a latent space where features from different cameras are correlated. Similarly, in~\cite{paper04}, the authors address the small-sample-size problem using shrinkage and smoothing techniques to better estimate the covariance matrices and Zhang et~\textit{al.}~\cite{paper82} collapse images of the same person in a single point in a discriminative null space. Prates and Schwartz~\cite{paper11} adapt Partial Least Squares (PLS) to a supervised Re-ID setting using prototypes to indirectly deal with camera transition. To tackle the nonlinearity of the data, Lisanti et~\textit{al.}~\cite{paper08} propose a kernel descriptor to encode person appearance and project the data into common subspace using Kernel Canonical Correlation Analysis (KCCA). Similarly, Kernel PLS~\cite{paper80} and Kernel HPCA~\cite{paper79} have been used to nonlinearly map data into a common subspace. As in the nonlinear subspace learning approaches, this work also  represents probe and gallery images nonlinearly by using a set of basis vectors. Differently, our basis vectors are composed of feature representation of each training samples. In fact, this feature representation can be obtained directly from feature descriptor or improved using projections to low-dimensional common subspace. %%Experimental results demonstrate that despite simple, this approach achieves higher recognition when compared with subspace learning methods.   
 
%% comentei o paragrafo sobre retrieval pois nao estah muito relacionado com que fazemos (pelo espaco reduzido que temos)
%%There are also efforts to close the gap between Re-ID and information retrieval approaches~\cite{paper06,paper16,paper20,paper13,paper14,paper15,paper88}. An et~\textit{al.}~\cite{paper06} address the person re-identification as a re-ranking problem using detection of soft-biometric attributes (e.g, hair size). Differently, Liu et~\textit{al.}~\cite{paper16} use human strong and/or weak feedback in a post-rank optimization framework. To reduce the human effort, unsupervised post-rank optimization methods have been proposed using least square regression models with manifold regularization~\cite{paper20} or content and context information~\cite{paper15}. In addition, some works have proposed to combine multiple experts at decision level using structural learning framework~\cite{paper14,paper13} or ranking aggregation~\cite{paper10}. More recently, the Deep Ranking~\cite{paper88} addresses Re-ID as a ranking problem by learning the feature representation and matching function in a unique deep learning framework. This work consists in pairwise matching between probe and gallery images without considering any attribute, user feedback or ranking information. Despite being useful, these information are not scalable to real-world scenarios where we can have dozens of cameras. Differently, the proposed~\emph{Kernel X-CRC} can be employed with reduced human effort to collect and annotate data.

Some works investigate the person re-identification problem using sparse or collaborative representations~\cite{paper47,paper17,paper55,paper56,paper57,paper58,paper75,paper76,paper92}. Lisanti et~\textit{al.}~\cite{paper17} propose an Iterative Sparse Ranking (ISR) method that iteratively applies SRC with adaptive weighting strategies until ranking all the gallery images. In~\cite{paper75,paper76,paper92}, the authors use CRC in the unsupervised multi-shot Re-ID scenario to compute the distance between probe and gallery images efficiently using both coding residuals and coefficients. Karanam et~\textit{al.}~\cite{paper47} explore the block structure in sparse coefficients to rank gallery images based on the reconstruction error. In~\cite{paper55,paper58}, the authors exploit dictionary learning and sparse coding in a unique framework. Differently, in~\cite{paper56,paper57}, the authors propose a local sparse representation method that uses SRC to represent interest points. \emph{Kernel X-CRC} has some key advantages when compared to these methods. For instance,~\emph{Kernel X-CRC} is a general method that does not assume a block structure in the coefficients representation as occurs in~\cite{paper47}. Differently from dictionary learning-based approaches~\cite{paper55,paper58}, this work represents probe and gallery images using training samples. More importantly, different from previous works~\cite{paper17,paper47,paper55,paper58}, we efficiently model the strong nonlinear transition of features between cameras achieving an analytical solution.  

Recently, some methods have employed multi-task learning in person re-identification~\cite{paper50,paper65,paper66}. Ma et~\textit{al.}~\cite{paper65} approach the person re-identification by transferring knowledge between the source domain with labeled image pairs and the target domain with unlabeled data using multi-task support vector ranking. In~\cite{paper66}, the authors avoid the over-fitting problem learning multiple Mahalanobis distance metrics in a multi-task framework. Su et~\textit{al.}~\cite{paper50} exploit the correlation between low-level features and attributes using a multi-task learning framework with low ranking embedding. The proposed~\emph{Kernel X-CRC} is fundamentally different from previous works because we employ multi-task to learn related collaborative coefficients from multiple regularized linear problems.

The method proposed in this work is related to joint sparse or collaborative representation methods. Such methods have been applied in multi-view face recognition~\cite{paper63,paper59}, hyperspectral image classification~\cite{paper62} and multi-modal biometrics recognition~\cite{paper64}. For instance, the works~\cite{paper62,paper54,paper59} use sparsity information to combine complementary features for classification. Differently, we are dealing with a single feature modality and the tasks correspond to the different camera-views. Shekhar~et~\textit{al.}~\cite{paper64} consider the correlation between multiple biometric information using sparse representation, while Zhang~et~\textit{al.}~\cite{paper63} employ a joint dynamic sparse representation to exploit the correlation between multiple views of same face image. However, they consider that testing and training images are available for different tasks (e.g., multiple views or modalities). Differently, Re-ID problem aims at predicting the subjects appearance at the target task (i.e., gallery camera) using the image from the source task (i.e., probe camera).
\section{Proposed Approach}

In the proposed \emph{Kernel X-CRC}, we use labeled training images at cameras A and B as columns of matrices $\textbf{D}_y$ and $\textbf{D}_x$, respectively (i.e., features belonging to the same subject are in corresponding columns in both matrix representations). Thus, these matrices encode the cross-view discriminative information that reflects in the learned collaborative representation coefficients ($\pmb{\alpha}_x$ and $\pmb{\alpha}_y$). For instance, when we describe two images of the same person captured by cameras A and B using $\textbf{D}_y$ and $\textbf{D}_x$, it is expected the representation coefficients to be more similar than when describing two different subjects. Therefore, we propose to use the similarity between these coefficients to indirectly compute the similarity between probe $\textbf{y}$ and the gallery-set~$\textbf{X}$. 

%\noindent\textbf{Notation.} 
We use the following notation in the description. Bold lower-case letters denote column vectors and bold upper-case letters denote matrices (e.g.,~\textbf{a} and~\textbf{A}, respectively). In this work, we deal with the single-shot scenario (i.e., there exist only one image taken from camera view A and one image taken from camera view B). We represent the~i\emph{th} image from camera A and B, as $\textbf{y}_i$ and $\textbf{x}_i~\in~\mathbb{R}^m$, respectively, where $m$ denotes the dimension of the feature space. Without loss of generality, we assume that $l$ testing images from camera A constitute the probe set $\textbf{Y} \in \mathbb{R}^{m \times l}$ and $l$ testing images from camera B represent the gallery set $\textbf{X} \in \mathbb{R}^{m \times l}$. Similarly, the set of all $n$ training images from camera A and B compose the matrices $\textbf{D}_y$ and $\textbf{D}_x \in \mathbb{R}^{m \times n}$, respectively. The collaborative representation coefficients are denoted by $\pmb{\alpha}_y, \pmb{\alpha}_x$, where $\pmb{\alpha} \in \mathbb{R}^n$, and~$\lambda \in \mathbb{R}$ is a scalar. We use $\phi(.)$ to denote a nonlinear mapping function of input variables to a feature space $\mathcal{F}$, i.e, $\phi : x_i \in \mathbb{R}^m \rightarrow \phi(x_i) \in \mathcal{F}$ and, $\pmb{\Phi}_x$ and $\pmb{\Phi}_y$ are the resulting matrices after nonlinearly mapping $\textbf{D}_x$ and $\textbf{D}_y$, respectively. In the following equations, we use the notation~\textbf{I} to indicate the identity matrix. 
%\vspace{\baselineskip}

%In the next section, we describe our proposed~\emph{Kernel X-CRC} and present an algorithm to efficiently compute collaborative representation coefficients $\pmb{\alpha}_x$ and $\pmb{\alpha}_y$. Notice that as we compute these coding vectors in a unique multi-task framework, they have to balance representativeness and similarity.

Considering as related tasks the representation of probe and gallery images using training images from their respective cameras, we propose to simultaneously estimate $\pmb{\alpha}_x$ and $\pmb{\alpha}_y$ in a multi-task collaborative representation framework. Thus, we aim at estimating the most similar coding vectors $\pmb{\alpha}_x$ and $\pmb{\alpha}_y$ that simultaneously describe probe and gallery subjects. To compute these coding vectors, we introduce a~\textit{similarity} term $\parallel\pmb{\alpha}_x - \pmb{\alpha}_y\parallel_2^2$ in our multi-task formulation that balances representativeness and similarity resulting in the following optimization problem  
\small
\begin{equation}
\begin{split}
\min_{\pmb{\alpha}_x,\pmb{\alpha}_y}\parallel\phi(\textbf{y}) - \pmb{\Phi}_y\pmb{\alpha}_y\parallel_2^2  +\parallel\phi(\textbf{x}) -\pmb{\Phi}_x \pmb{\alpha}_x\parallel_2^2 \\
+\lambda\parallel\pmb{\alpha}_y\parallel_2^2 + \lambda\parallel\pmb{\alpha}_x\parallel_2^2 + \parallel\pmb{\alpha}_y -\pmb{\alpha}_x\parallel_2^2,
\label{eq:eq05}
\end{split}
\end{equation}
\normalsize
that we analytically derived with respect to $\pmb{\alpha}_y$ and $\pmb{\alpha}_x$ obtaining 
\begin{equation}
\pmb{\alpha}_y = \textbf{P}_y^{-1}\pmb{\alpha}_x + \textbf{P}_y^{-1}\pmb{\Phi}_y^{\top}\phi(\textbf{y})
\label{eq:09}
\end{equation}
and
\begin{equation}
\pmb{\alpha}_x = \textbf{P}_x^{-1}\pmb{\alpha}_y + \textbf{P}_x^{-1}\pmb{\Phi}_x^{\top}\phi(\textbf{x}),
\label{eq:10}
\end{equation}
where projections matrices~$\textbf{P}_y$ and~$\textbf{P}_x$ are given by
\begin{equation}
\begin{split}
\textbf{P}_y = \pmb{\Phi}_y^{\top}\pmb{\Phi}_y + \lambda\textbf{I} \text{ and }
\textbf{P}_x = \pmb{\Phi}_x^{\top}\pmb{\Phi}_x + \lambda\textbf{I}.
\end{split}
\label{eq:eq07}
\end{equation}
Note that Equations~\ref{eq:09} and~\ref{eq:10} are interdependent. Therefore, replacing~$\pmb{\alpha}_x$ for its corresponding equation (Eq.~\ref{eq:10}) and isolating~$\pmb{\alpha}_y$, we obtain
\begin{equation}
\pmb{\alpha}_y = \textbf{Q}^{-1}\textbf{P}_y^{-1}\textbf{P}_x^{-1}\pmb{\Phi}_x^{\top}\phi(\textbf{x}) + \textbf{Q}^{-1}\textbf{P}_y^{-1}\pmb{\Phi}_y^{\top}\phi(\textbf{y})
\label{eq:12}
\end{equation}
with projection matrix \textbf{Q} corresponding to
\begin{equation}
\textbf{Q} = \textbf{I} - \textbf{P}_y^{-1}\textbf{P}_x^{-1}.
\label{eq:eq11}
\end{equation}
Similarly, we can compute the coding vector $\pmb{\alpha}_x$ as
\begin{equation}
\pmb{\alpha}_x = \textbf{W}^{-1}\textbf{P}_x^{-1}\textbf{P}_y^{-1}\pmb{\Phi}_y^{\top}\phi(\textbf{y}) + \textbf{W}^{-1}\textbf{P}_x^{-1}\pmb{\Phi}_x^{\top}\phi(\textbf{x})
\label{eq:eq:novel_eq}
\end{equation}
with \textbf{W} computed as
\begin{equation}
\textbf{W} = \textbf{I} - \textbf{P}_x^{-1}\textbf{P}_y^{-1}.
\label{eq:eq112}
\end{equation}

To avoid explicitly mapping of data to a high-dimensional space, we can use the ``kernel trick'' substituting cross-product by $\textbf{K} = \pmb\Phi\pmb\Phi^\top$, where $\textbf{K} \in \mathbb{R}^{n \times n}$ is the \textit{kernel Gram matrix}. Particularly, we define the \textit{kernel Gram matrices} $\textbf{K}_x$ and $\textbf{K}_y \in \mathbb{R}^{n \times n}$ to represent the cross-product $\pmb{\Phi}_x^{\top}\pmb{\Phi}_x$ and $\pmb{\Phi}_y^{\top}\pmb{\Phi}_y$, respectively. Furthermore, we define $\pmb{\Phi}_x^{\top}\phi(\textbf{x})$ as the computation of kernel function between $\textbf{x}$ and all vectors $\hat{\textbf{x}} \in \pmb{\Phi}_x$. Identically, $\pmb{\Phi}_y^{\top}\phi(\textbf{y})$ denotes the kernel function applied in $\textbf{y}$ and all vectors $\hat{\textbf{y}} \in \pmb{\Phi}_y$.  Then, the similarity between a pair of probe~\textbf{y} and gallery~\textbf{x} is computed by the similarity between $\pmb{\alpha}_x$ and $\pmb{\alpha}_y$, as described in Algorithm~\ref{alg:alg01}.

\begin{algorithm}
\DontPrintSemicolon
\SetAlgoNoLine
\SetKwInOut{Input}{input}\SetKwInOut{Output}{output}
\Input{Kernel matrices ($\textbf{K}_x$ and $\textbf{K}_y$)}
\Output{Ranking list of gallery images \textbf{R}}
\BlankLine
$\textit{Compute $\textbf{P}_x$ and $\textbf{P}_y$ matrices using Equation~\ref{eq:eq07}}$ \;
$\textit{Compute $\textbf{Q}$ and $\textbf{W}$ using Equations~\ref{eq:eq11} and~\ref{eq:eq112}}$\;
$\textit{Pre-compute:}$\

$\pmb{\beta}_x^{x} \leftarrow \textbf{W}^{-1}\textbf{P}_x^{-1}$\;
$\pmb{\beta}_x^{y} \leftarrow \textbf{W}^{-1}\textbf{P}_x^{-1}\textbf{P}_y^{-1}$\;
$\pmb{\beta}_y^{y} \leftarrow \textbf{Q}^{-1}\textbf{P}_y^{-1}$\;
$\pmb{\beta}_y^{x} \leftarrow \textbf{Q}^{-1}\textbf{P}_y^{-1}\textbf{P}_x^{-1}$\;
\For{$\textbf{y}_j \in \textbf{Y}$}{
	\For{$\textbf{x}_i \in \textbf{X}$}{
		  $\pmb{\alpha}_x \leftarrow \pmb{\beta}_x^{x}\pmb{\Phi}_x^{\top}\phi(\textbf{x}_i) +\pmb{\beta}_x^{y}\pmb{\Phi}_y^{\top}\phi(\textbf{y}_j)$ \;
		  
		  $\pmb{\alpha}_y \leftarrow \pmb{\beta}_y^{x}\pmb{\Phi}_x^{\top}\phi(\textbf{x}_i) +\pmb{\beta}_y^{y}\pmb{\Phi}_y^{\top}\phi(\textbf{y}_j)$ \;
		  
		  $ sim(i) \leftarrow \frac{\pmb{\alpha}_x^{\top}\pmb{\alpha}_y}{\parallel \pmb{\alpha}_x \parallel \parallel \pmb{\alpha}_y \parallel} $ \;
	  }
	  $\textbf{R}_j \leftarrow sort(sim,descend)$\;
 }
 \Return{\textbf{R}}\;
 \caption{\emph{Kernel Cross-View Collaborative Representation based Classification (Kernel X-CRC)}.}
  \label{alg:alg01}
\end{algorithm}

Due the multi-task learning framework, a pair of probe (\textbf{y}) and gallery images (\textbf{x}) will compute $\pmb{\alpha}_y$ and $\pmb{\alpha}_x$ that balances the representativeness in each camera with the similarity between coding vectors. This balance will only result in a similar coding vector if~\textbf{x} corresponds to the respective gallery image of \textbf{y}. For instance, if~\textbf{x} is dissimilar when compared to~\textbf{y}, similar coding vectors will not be obtained since they should result in poor representativeness (i.e., high reconstruction error) in both cameras. %In the next section, we present some experimental results that corroborate with our assumption.
\section{Experimental Results}

%\begin{figure*}[!tbp]
%  \centering
%  \subfloat[PRID450S]{\includegraphics[width=0.48\textwidth]{figures/prid450s_dataset_examples}}
%\hspace{2mm}
%  \subfloat[VIPER]{\includegraphics[width=0.48\textwidth]{figures/viper_dataset_examples}}
%   \caption{Different persons (columns) captured by two surveillance cameras (rows) in PRID450S (left image) and VIPeR (right image) datasets.}
%  \label{fig:prid450s_viper}
%\end{figure*}

In this section, we perform a comprehensive evaluation of the proposed~\emph{Kernel X-CRC} assessing the effect of different strategies in the experimental results (Section~\ref{sec:xcrc:aspects}) and providing a broad comparison with other approaches in the state-of-the-art in four datasets (Section~\ref{sec:stateoftheart}).

%\subsection{Datasets and Experimental Setup}
%\label{sec:datasets}

\vspace{1mm}

\noindent\textbf{Datasets.} To perform our experiments, we consider four challenging datasets. The \emph{PRID 450S Dataset}\footnote{Available at: $\text{https://lrs.icg.tugraz.at/download.php}$}~\cite{paper74} consists of 450 images pairs of pedestrians captured by two non-overlapping cameras. The main challenges are related to changes in viewpoint, pose as well as significant differences in background and illumination. The \emph{VIPeR Dataset}\footnote{Available at: https://vision.soe.ucsc.edu/projects}~\cite{paper68} contains 632 labelled image pairs captured by two different outdoor cameras located in an academic environment. Each subject appears once in each camera and most of the image pairs show viewpoint change larger than 90 degrees, making it a very challenging dataset. The \emph{CUHK01 Dataset}\footnote{Available at:http://www.ee.cuhk.edu.hk/rzhao/}~\cite{paper67} captures two disjoint camera-view images for each person in a campus environment, containing 971 persons, each of which has two images from each camera-view (all the images are normalized to $160\times60$ pixels for evaluations). The \emph{GRID dataset}\footnote{Available at: http://personal.ie.cuhk.edu.hk/ \textasciitilde ccloy}~\cite{paper93} contains 250 image pairs (single-shot) captured by eight disjoint surveillance cameras in a busy underground station generating different poses and poor illumination conditions. In addition, different from the other datasets evaluated, it introduces 775 individuals in gallery-set without correct matching in the probe-set (distractors) that drastically impacts in the results.

\vspace{1mm}

\noindent\textbf{Experimental Setup.}
%\noindent\emph{Experimental Setup.}
As in the majority of the works, we randomly partition the datasets into training and testing subsets with an equal number of individuals. However, due to the odd number of subjects, in CUHK01, we split the 971 individuals into 485 persons for training and the remaining 486 for testing. Furthermore, we also adapt CUHK01 to the single-shot scenario by randomly selecting one image of the same person in each camera, similarly to~\cite{paper81}.

To set $\lambda$, the only parameter of the proposed~\emph{Kernel X-CRC}, we use a single partition for each dataset. It differs from the multiple partitions commonly used in literature and avoids overfitting the parameter to the data. We also use this single partition to define exponential $\chi^2$ and RBF as kernel functions employed when dealing with the low-dimensional and original descriptors, respectively.

We report the average of results obtained from 10 trials, a common procedure to achieve more stable results. The results are reported using the~\emph{rank-k} matching rate, which consists on the percentage of individuals correctly identified when considering the~\textit{top-k} ranking positions, a widely employed metric to compare Re-ID approaches. We present the evaluated approaches in tables using an ascending order of reported~\emph{rank-1} matching rate.

%\begin{figure*}[!tbp]
%  \centering
%  \subfloat[CUHK01]{\includegraphics[width=0.475\textwidth]{figures/cuhk01_dataset_examples}}
% \hspace{2mm}
%   \subfloat[GRID]{\includegraphics[width=0.48\textwidth]{figures/grid_dataset}}
%   \caption{Examples of individuals (columns) captured by different surveillance cameras (rows) in CUHK01 (left image) and GRID (right image) datasets.}
%  \label{fig:grid_cuhk01}
%\end{figure*}

\subsection{Kernel X-CRC Evaluation}
\label{sec:xcrc:aspects}

%Firstly, we show that the proposed method reaches interesting results when working with feature descriptors in the original space and that these results increase when the feature descriptor is more robust to different camera conditions (Section~\ref{sec:features}) . Then, Section~\ref{sec:features_xqda} demonstrates a great improvement when using  XQDA~\cite{paper45} to learn a common subspace that better handles with the camera transition problem. Afterwards, we execute experiments to determine the contribution of~\emph{Kernel X-CRC} when computed in the common subspace (Section~\ref{sec:matching_models}). Specifically, we show that~\emph{Kernel X-CRC} greatly improves the obtained experimental results when compared to the traditional approach of matching probe and gallery images in the common subspace using simple cosine or Mahalanobis metric functions.

In this section, we use the VIPeR dataset to evaluate the performance of~\emph{Kernel X-CRC} according to different aspects: feature descriptors in the original space, performance when operating in a common subspace, contribution of~\emph{Kernel X-CRC} when computed in the common subspace, and the impact of the different choices that resulted in the proposed~\emph{Kernel X-CRC}.

%First, we show that the Kernel X-CRC reaches interesting results when working with feature descriptors in the original space. Second, we show that a great improvement can be achieved when XQDA~\cite{paper45} is employed to learn a common subspace that better handles with the camera transition problem.

\vspace{1mm}

\noindent\textbf{Feature Descriptor Evaluation.}
%\label{sec:features}
This experiment assesses the performance of~\emph{Kernel X-CRC} using feature descriptors widely employed  in the literature. We used the descriptors proposed by Zheng et~\emph{al.}~\cite{paper73} and Koestinger et~\emph{al.}~\cite{paper71} that are simple combination of color histograms and texture information extracted from local patches. Due to their limited capability of dealing with the camera transition problem, we also considered descriptors that better handle this problem, such as LOMO~\cite{paper45}, WHOS~\cite{paper08}, GoG~\cite{paper81} and LOMO+CNN~\cite{paper72}.

%. LOMO~\cite{paper45}  uses horizontal max-pooling, Retinex transformation and multiple scales to handle with  viewpoints, different illumination conditions and spatial misalignments. Differently, WHOS~\cite{paper08} handles the camera transition by using overlapping stripes, while GoG~\cite{paper81} and LOMO+CNN~\cite{paper72} compute feature descriptors hierarchically. 

Table~\ref{table:features} shows that the~\emph{Kernel X-CRC} is able to obtain more accurate results when using a better feature representation. For instance,~\emph{Kernel X-CRC} reached $45.5\%$ of~\emph{rank-1} using LOMO+CNN~\cite{paper72}, which is comparable with state-of-the-art approaches. Due to the superior performance of LOMO~\cite{paper45}, WHOS~\cite{paper08}, GoG~\cite{paper81} and LOMO+CNN~\cite{paper72}, we will focus on these feature descriptors in the following experiments.

\begin{table}[!htb]
\tiny
\centering
		\resizebox{1\columnwidth}{!}{
		\begin{tabular}{c|c|c|c|c|c}
			\hline
			\multirow{2}{*}{Feature Descriptor} & \multicolumn{5}{c}{Viper (p=316)} \\ \cline{2-6}
			& r = 1 & r = 5 & r = 10 & r = 20 & r = 30 \\ \hline 
			Zheng et~\emph{al.}~\cite{paper73} & 21.8 & 52.2 & 68.2 & 83.3 & 89.3 \\ 
			Koestinger et~\emph{al.}~\cite{paper71} & 22.5 & 53.2 & 68.8 & 82.6 & 89.2 \\ 
			LOMO~\cite{paper45} & 37.8  & 72.1 & 85.4 & 94.2 & 97.0\\ 
			WHOS~\cite{paper08} & 42.9 & 77.4 & 88.3 & 94.5 & 97.0\\ 
			GoG~\cite{paper81} & 45.0 & 78.1 & 88.7 & \bf{96.1} & \bf{98.3}\\ 
			LOMO+CNN~\cite{paper72} & \bf{45.5} & \bf{78.9} & \bf{88.9} & 95.9 & 98.1 \\ \hline
		\end{tabular}
		}
		\caption{Feature descriptor evaluation on the VIPeR dataset.}
		\label{table:features}
\end{table}

\vspace{1mm}

\noindent\textbf{Subspace Evaluation.}
%\label{sec:features_xqda}
Based on the results obtained using different feature descriptors (see Table~\ref{table:features}), one might hypothesize that better results can be achieved by improving the feature representation. A straightforward approach for achieving  a better feature representation consists in concatenating complementary feature descriptors. However, as we solve regularized linear models for each pair of probe and gallery images, it will result in a prohibitive computational cost. An alternative is to compute a low-dimensional subspace that maintains the computational cost acceptable yet improves the results. Therefore, we project the data onto a common subspace that handles the camera transition problem before matching the probe with gallery images using the proposed~\emph{Kernel X-CRC}.

Table~\ref{table:features_xqda} presents the experimental results using~\emph{Kernel X-CRC} in a low-dimensional feature representation computed using XQDA~\cite{paper45}. The results were improved for all feature descriptors when compared to those shown in Table~\ref{table:features}. It is important to highlight that we employed XQDA due to outperforming results reported in literature yet, other methods to estimate common subspaces could be employed, instead.

%its online available code and the outperforming results reported in literature. %However, the proposed~\emph{Kernel X-CRC} is a general method that can be employed with the original feature representation or any other common subspace learning method.

%Note that the results were improved for all feature descriptors by using the low-dimensional representation. %For instance, GoG~\cite{paper81} improves more than 6.0 percentage points at~\emph{rank-1}. 
%Therefore, in the remaining experiments, we apply~\emph{Kernel X-CRC} in the feature representation learned using XQDA and considering the GoG feature descriptor.
\begin{table}[!htb]
\tiny
\centering
		\resizebox{1\columnwidth}{!}{
		\begin{tabular}{c|c|c|c|c|c}
			\hline
			\multirow{2}{*}{Feature Descriptor} & \multicolumn{5}{c}{Viper (p=316)} \\ \cline{2-6}
			& r = 1 & r = 5 & r = 10 & r = 20 & r = 30 \\ \hline 
			%Koestinger et~\emph{al.}~\cite{paper71} & 20.1 & 42.0 & 54.8 & 68.7 & 76.6\\ \hline
			%Zheng et~\emph{al.}~\cite{paper73} & 24.2 & 53.1 & 67.8 & 80.7 & 87.6\\ \hline
			LOMO~\cite{paper45} & 41.7  & 72.2 & 84.0 & 93.6 & 96.3\\ 
			WHOS~\cite{paper08} & 43.3 & 73.6 & 84.8 & 92.7 & 96.3\\ 
			LOMO+CNN~\cite{paper72} & 47.1 & 77.3 & 88.6 & 95.9 & 98.2\\ 
			GoG~\cite{paper81} & \bf{51.6} & \bf{80.8} & \bf{89.4} & \bf{95.3} & \bf{97.4}\\ \hline
		\end{tabular}
		}
		\caption{Subspace evaluation on the VIPeR dataset.}
		\label{table:features_xqda}
\end{table}

\vspace{1mm}

\noindent\textbf{Metric Function Evaluation.}
According to the previous experiment, the employment of~\emph{Kernel X-CRC} in a low-dimensional feature space learned using XQDA improved the results. However, it is difficult to define whether the improvement gain are due to~\emph{Kernel X-CRC} or to the better representation learned using XQDA. Therefore, to highlight the contribution of~\emph{Kernel X-CRC}, we compare the proposed method with traditional metric functions to match probe and gallery images in the learned common subspace. Specifically, we compare~\emph{Kernel X-CRC} with cosine and Mahalanobis distances and KISSME metric~\cite{paper71}. % using the~\emph{rank-1} matching rate. %, as shown in Table~\ref{table:xqda_matching}.  

\begin{table}[!htb]
\tiny
\centering
		\resizebox{1\columnwidth}{!}{
		\begin{tabular}{c|c|c|c|c}
			\hline
			\multirow{2}{*}{Function} & \multicolumn{4}{c}{Viper (p=316)} \\ \cline{2-5}
			& WHOS~\cite{paper08} & LOMO~\cite{paper45} & GoG~\cite{paper81} & LOMO+CNN~\cite{paper72} \\ \hline 
			Cosine & 33.7 & 33.5 & 47.1 & 40.4 \\ 
			Mahalanobis  & 34.1 & 38.3 & 46.5 & 43.2 \\ 
			KISSME  & 34.0 & 35.1 & 43.2 & 41.0 \\
			\bf{Kernel X-CRC}  & \bf{43.3} & \bf{41.7} & \bf{51.6} & \bf{47.1} \\ \hline
		\end{tabular}
		}
		\caption{Metric functions evaluation on the VIPeR dataset.}
		\label{table:xqda_matching}
\end{table}

According to Table~\ref{table:xqda_matching},  even though reasonable results were achieved  in the learned subspace when using the traditional metric functions (e.g., cosine and Mahalanobis distance), %due to  (this behavior occurs because XQDA learns projections to the common subspace that keep images of the same person closer than different persons)
the \emph{Kernel X-CRC} achieved greater improvements (5.3 percentage points, on average), when compared to the metric function with the second highest~\emph{rank-1} matching rate.
% the improvement achieved by the~\emph{Kernel X-CRC}, compared to the metric function with second highest~\emph{rank-1} rate, was 5.3 percentage points. 
We attribute this performance gain to the computation of specific coding vectors for each pair of probe and gallery images using a nonlinear model. On the other hand, cosine and Mahalanobis are fixed distances, and the KISSME~\cite{paper71} is a global distance learned using the entire training set.

% interesting results were obtained  in the learned subspace even when using simple cosine and Mahalanobis distance, which are traditional metric functions used in literature~\cite{paper45,paper81}. This behavior occurs because XQDA learns projections to the common subspace that keep images of the same person closer than different persons. More importantly, Table~\ref{table:xqda_matching} shows that~\emph{Kernel X-CRC} was able to achieve a great improvement when compared to traditional approaches. % For instance, when using GoG~\cite{paper81}, we reached 51.6\% of~\emph{rank-1} with the proposed method, 4.5 percentage points higher than the result achieved using cosine distance. 

Due the improved results, the remaining experiments will consider the GoG~\cite{paper81} descriptor in the low-dimensional representation computed using XQDA.

\vspace{1mm}

\noindent\textbf{Baseline Approaches.}
%\label{sec:approaches}
This experiment analyzes the impact of different choices that resulted in the~\emph{Kernel X-CRC} model. We evaluate unsupervised methods (SRC~\cite{paper53} and CRC~\cite{paper53}) and supervised methods with and without considering multi-task and kernel extensions. We also compare~\emph{Kernel X-CRC} with the model presented in Equation~\ref{eq:eq02}, which we named \textit{Cross-View Collaborative Representation classification} (C\textsuperscript{2}RC), and with a straightforward nonlinear extension of C\textsuperscript{2}RC, referred to as \emph{Kernel C\textsuperscript{2}RC}.

%Table~\ref{table:approaches} presents the results considering different coding representation strategies using the GoG feature representation in the low-dimensional space computed employing XQDA. 

According to Table~\ref{table:approaches}, SRC~\cite{paper53} and CRC~\cite{paper53} achieved the lowest results as they directly match probe and gallery images based on the reconstruction error without considering the training samples. Differently, C\textsuperscript{2}RC  and Kernel C\textsuperscript{2}RC improved the results for all ranking positions by using the training samples to relate features with coding vectors that are employed to indirectly match probe and gallery images. Furthermore, due to the nonlinear modeling, the Kernel C\textsuperscript{2}RC reached results even better than its linear counterpart, the C\textsuperscript{2}RC.

\begin{table}[!htb]
\tiny
\centering
		\resizebox{1\columnwidth}{!}{
		\begin{tabular}{c|c|c|c|c|c}
			\hline
			\multirow{2}{*}{Approach} & \multicolumn{5}{c}{Viper (p=316)} \\ \cline{2-6}
			& r = 1 & r = 5 & r = 10 & r = 20 & r = 30 \\ \hline 
			SRC & 20.6 & 37.3 & 49.0 & 63.0 & 68.7 \\ 
			CRC & 22.5 & 38.6 & 48.1  & 60.1 & 67.4 \\
			$$C\textsuperscript{2}RC$$  & 49.9 & 78.6 & 87.7 & 94.4 & 96.5\\ 
			X-CRC & 50.5 & 79.3 & 88.7 & 94.5 & 97.0\\ 
			Kernel $$C\textsuperscript{2}RC$$  & 51.0 & 79.4 & 88.9 & 94.8 & 96.9\\ 
			\bf{Kernel X-CRC} & \bf{51.6} & \bf{80.8} & \bf{89.4} & \bf{95.3} & \bf{97.4}\\ \hline
		\end{tabular}
		}
		\caption{Results of the baseline approaches on the VIPeR dataset.}
		\label{table:approaches}		
\end{table}

The employment of the X-CRC and~\emph{Kernel X-CRC} models achieved improved results when compared to their counterparts C\textsuperscript{2}RC and Kernel C\textsuperscript{2}RC, respectively. We attribute this gain to the multi-task learning framework that forces the coding vectors to be simultaneously representative in each camera and similar between different cameras.

%According to the results in Table~\ref{table:approaches}, the X-CRC and~\emph{Kernel X-CRC} models achieved improved results when compared to its counterparts C\textsuperscript{2}RC and Kernel C\textsuperscript{2}RC, respectively. We relate this performance gain to the multi-task learning framework that forces the coding vectors to be simultaneously representative in each camera and similar between different cameras.

\subsection{State-of-the-art Comparisons}
\label{sec:stateoftheart}

In this section, we compare the proposed approach with a large number of state-of-the-art methods on the VIPeR, PRID450S, CUHK01 and GRID datasets.

\vspace{1mm}

\noindent\textbf{VIPeR dataset.}
%\label{sec:viper}
%In this section, we compare the proposed~\emph{Kernel X-CRC} with state-of-the-art methods from literature in VIPeR dataset. 
Table~\ref{table:viper} presents the matching rates for different methods, including  approaches based on metric learning~\cite{paper85,paper84,paper83}, common subspace learning~\cite{paper80,paper08,paper79,paper45,paper82,paper90,paper81,paper10} and deep learning~\cite{paper89,paper88,paper72}. According to the results, the proposed method greatly outperforms most of the approaches. For instance, the~\emph{Kernel X-CRC} reached 51.6\% of~\emph{rank-1}, while Wu et~\emph{al.}~\cite{paper72} achieved 51.1\% combining LOMO and feature descriptors based on deep learning architectures. We believe that our simple approach can outperform more complex methods (e.g., deep learning based approaches) because it learns a coding representation specific for a pair of probe and gallery images, while other methods attempt to learn a matching model considering a small training set, which is prone to overfitting. Similarly to our work, Zhang et~\emph{al.}~\cite{paper91} also employs a specific model for each pair of probe and gallery images. However, their models are obtained using a mapping function to relate feature descriptors to model parameters, which is very challenging for small datasets such as VIPeR.

\begin{table}[!htb]
\tiny
\centering
		\resizebox{1\columnwidth}{!}{
		\begin{tabular}{c|c|c|c|c|c}
			\hline
			\multirow{2}{*}{Method} & \multicolumn{5}{c}{Viper (p=316)} \\ \cline{2-6}
			& r = 1 & r = 5 & r = 10 & r = 20 & r = 30 \\ \hline 
			Prates and Schwartz~\cite{paper10}& 32.9 & 62.3 & 78.7 & 87.8 & 91.6 \\
			KPLS ModeA~\cite{paper80} & 35.8 & 69.1 & 80.8 & 89.9 & 93.8 \\
			KCCA~\cite{paper08} & 37.0 & - & 85.0 & 93.0 & - \\
			X-KPLS~\cite{paper80} & 38.4 & 73.0 & 85.2 & 93.4 & 94.5 \\ 
			Deep Ranking~\cite{paper88} & 38.4 & 69.2 & 81.3 & 90.4 & 94.1 \\ 
			KISSME & 39.2 & 71.8 & 81.3 & 92.4 & 94.9\\ 			
			Kernel HPCA~\cite{paper79}  & 39.4 & 73.0 & 85.1 & 93.5 & 96.1 \\ 
			LOMO + XQDA~\cite{paper45} & 40.0 & 68.0 & 80.5 & 91.1 & 95.5 \\
			WARCA~\cite{paper84} & 40.2 & 68.2 & 80.7 & 91.1 & - \\ 
			NLML~\cite{paper85} & 42.3 & 71.0 & 85.2 & 94.2 & - \\ 	
			Null Space~\cite{paper82} & 42.3 & 71.5 & 82.9 & 92.1 & - \\ 		
			Zhang et~\emph{al.}~\cite{paper91} & 42.7 & - & 84.3 & 91.9 & - \\ 
			Mirror + KMFA~\cite{paper90} & 43.0 & 75.8 & 87.3 & 94.8 & - \\
			Sakrapee et~\emph{al.}~\cite{paper14} & 45.9 & 77.5 & 88.9 & 95.8 & - \\
			MultiCNN~\cite{paper89} & 47.8 & 74.7 & 84.8 & 91.1 & 94.3\\
			GoG + XQDA~\cite{paper81}  & 49.7 & 79.7 & 88.7 & 94.5 & -\\
			Wu et~\emph{al.}~\cite{paper72} & 51.1 & 81.0 & 91.4 & 96.9 & - \\  
			SCSP~\cite{paper83}  & \bf{53.5} & \bf{82.6} & \bf{91.5} & \bf{96.6} & - \\  \hline
			\bf{Kernel X-CRC} & 51.6 & 80.8 & 89.4 & 95.3 & \bf{97.4}\\ \hline
		\end{tabular}
		}
		\caption{Top ranked approaches on the VIPer dataset.}		
		\label{table:viper}		
\end{table}

To the best of our knowledge, SCSP~\cite{paper83} is the unique method with improved results when compared to the proposed~\emph{Kernel X-CRC}. We credit the better results to the combination of global and local matching models. However, learning how to constraint the matching of local regions in images obtained from different camera-views is a very challenge task, mainly in more realistic datasets.

\vspace{1mm}

\noindent\textbf{PRID450S dataset.}
%\label{sec:prid450S}
%In this section, we compare the proposed~\emph{Kernel X-CRC} with methods from literature in PRID450S, as shown in Table~\ref{table:prid450s}. 
According to the results shown in Table~\ref{table:prid450s} and similarly to the results in VIPeR dataset (Table~\ref{table:viper}), we achieved improved results when compared to distance metric learning~\cite{paper84}, subspace learning~\cite{paper90,paper45,paper80,paper79,paper81,paper10}, deep learning~\cite{paper72} and other approaches. For instance, using the~\emph{Kernel X-CRC}, we reached higher matching rates for all ranking positions when compared to GoG + XQDA~\cite{paper81}. That can be explained by the use of the same low-dimensional representation in the common subspace to nonlinearly compute coding vectors specific to each pair of probe and gallery images, while they compute a simple cosine distance. To the best of our knowledge, we achieved the best reported results in literature for PRID450S. We attribute the improvement to the better representation obtained using GoG descriptor with XQDA~\cite{paper81} and the nonlinear computation of coding vectors by the~\emph{Kernel X-CRC}.

\begin{table}[!htb]
\tiny
\centering
		\resizebox{1\columnwidth}{!}{
		\begin{tabular}{c|c|c|c|c|c}
			\hline
			\multirow{2}{*}{Method} & \multicolumn{5}{c}{PRID450S (p=225)} \\ \cline{2-6}
			& r = 1 & r = 5 & r = 10 & r = 20 & r = 30 \\ \hline 
			WARCA~\cite{paper84} & 24.6 & 55.5 & 70.3 & 85.0 & 92.0 \\ 
			Prates and Schwartz~\cite{paper10}& 29.3 & 52.5 & 63.1 & 75.0 & 82.1 \\
			SCSP~\cite{paper83}  & 44.4 & 71.6 & 82.2 & 89.8 & 93.3 \\
			KPLS ModeA~\cite{paper80} & 51.5 & 78.6 & 87.0 & 93.7 & 96.0 \\			
			Kernel HPCA~\cite{paper79}  & 52.8 & 80.9 & 89.0 & 95.1 & 97.2 \\ 
			X-KPLS~\cite{paper80} & 52.8 & 82.1 & 90.0 & 95.4 & 97.3 \\	
			KISSME & 56.8 & 82.7 & 89.4 & 94.6 & 97.1\\ 
			Mirror + KMFA~\cite{paper90} & 55.4 & 79.3 & 87.8 & 91.6 & - \\
			Zhang et~\emph{al.}~\cite{paper91} & 60.5 & - & 88.6 & 93.6 & - \\ 
			LOMO + XQDA~\cite{paper45} & 61.4 & - & 90.8 & 95.3 & - \\
			Wu et~\emph{al.}~\cite{paper72} & 66.6 & 86.8 & 92.8 & 96.9 & - \\ 
			GoG + XQDA~\cite{paper81}  & 68.4 & 88.8 & 94.5 & 97.8 & -\\	\hline
			\bf{Kernel X-CRC} &\bf{68.8} & \bf{91.2} & \bf{95.9} & \bf{98.4} & \bf{99.0}\\ \hline
		\end{tabular}
		}
	\caption{Top ranked approaches on the PRID450S dataset.}
		\label{table:prid450s}	
\end{table}

\vspace{1mm}

\noindent\textbf{CUHK01 dataset.}
%\label{sec:cuhk01}
Table~\ref{table:cuhk01} presents the matching rates for different ranking positions of state-of-the-art methods that addressed person ReID on the CUHK01 dataset using the single-shot setting. Notice that results from LOMO + XQDA~\cite{paper45} and Zhang et~\emph{al.}~\cite{paper91} are not included since they focus on the multi-shot scenario. According to the results, the proposed~\emph{Kernel X-CRC} reaches better results than most of the metric learning and subspace learning approaches. As matter of fact, we reached the second best~\emph{rank-1} matching rate reported in literature, only surpassed by WARCA~\cite{paper84}, which is a nonlinear metric model. Regarding the higher ranking positions (e.g., r=10, 20 and 30), the best results are achieved using MultiCNN~\cite{paper89}, which consists in multiple deep learning architectures to combine local and global features. It is important to highlight that both methods WARCA and MultiCNN are very sensitive to the number of training samples available, which justifies the reduced matching rates reported in small datasets. Differently, the~\emph{Kernel X-CRC} remains within the top two best~\emph{rank-1} approaches for all evaluated datasets.
\begin{table}[!htb]
\tiny
\centering
		\resizebox{1\columnwidth}{!}{
		\begin{tabular}{c|c|c|c|c|c}
			\hline
			\multirow{2}{*}{Method} & \multicolumn{5}{c}{CUHK01 (p=485)} \\ \cline{2-6}
			& r = 1 & r = 5 & r = 10 & r = 20 & r = 30 \\ \hline
			KPLS ModeA & 38.3 & 66.8 & 77.7 & 86.8 & 90.5 \\
			Mirror + KMFA~\cite{paper90} & 40.4 & 64.6 & 75.3 & 84.1 & - \\
			Kernel HPCA  & 44.3 & 73.2 & 82.7 & 90.1 & 93.4\\ 
			X-KPLS & 46.2 & 74.0 & 84.3  & 91.3 & 94.0  \\	
			KISSME & 49.6 & 74.7 & 83.8 & 91.2 & 94.3\\ 
			Sakrapee et~\emph{al.}~\cite{paper14} & 53.4 & 76.4 & 84.4 & 90.5 & - \\
			MultiCNN~\cite{paper89} & 53.7 & 84.3 & \bf{91.0} & \bf{96.3} & \bf{98.3}\\
			GoG + XQDA~\cite{paper81}  & 57.8 & 79.1 & 86.2 & 92.1 & -\\
			WARCA~\cite{paper84} & \bf{65.6} & \bf{85.3} & 90.5 & 95.0 & -\\  \hline
			%LOMO + XQDA~\cite{paper45} & 63.2 & 84.0 & 90.0 & 93.7 & - \\
			%Zhang et~\emph{al.}~\cite{paper91} & \bf{66.0} & - & - & - & - \\ \hline \hline
			\bf{Kernel X-CRC} & 61.2 & 80.9 & 87.3 & 93.2 & 95.6 \\ \hline
		\end{tabular}
		}
		\caption{Top ranked approaches on the CUHK01 dataset.}
		\label{table:cuhk01}
\end{table}

\vspace{1mm}

\noindent\textbf{GRID dataset.}
%\label{sec:grid}
%
%In this section, we compare the obtained experimental results using~\emph{Kernel X-CRC} and other approaches from literature. 
Since it is a very challenging dataset that considers real-world problems (e.g., distractors), there are few works with reported experimental results on GRID dataset. Table~\ref{table:grid} presents these approaches and their respective matching rates for different ranking positions. Based on the results, we observe that GoG captures a better feature representation than the LOMO descriptors. For instance, when comparing both in the subspace learned using XQDA, GoG achieves an improvement of more than 8.0 percentage points for~\emph{rank-1} matching rate. Furthermore, GoG + XQDA~\cite{paper81} holds higher matching rates than different approaches in literature based on metric learning~\cite{paper85}, local and global matching~\cite{paper83} and the sample-specific matching models~\cite{paper91}. More importantly,~\emph{Kernel X-CRC} presents the highest~\emph{rank-1} matching rate when compared to these approaches, demonstrating the advantage of the nonlinear matching and its robustness to distractors.

\begin{table}[!htb]
\tiny
\centering
		
		\resizebox{1\columnwidth}{!}{
		\begin{tabular}{c|c|c|c|c|c}
			\hline
			\multirow{2}{*}{Method} & \multicolumn{5}{c}{GRID (p=125)} \\ \cline{2-6}
			& r = 1 & r = 5 & r = 10 & r = 20 & r = 30 \\ \hline
			LOMO + XQDA~\cite{paper45} & 16.6 & 33.8 & 41.8 & 52.4 & - \\
			Zhang et~\emph{al.}~\cite{paper91} & 22.4 & - & 51.3 & 61.2 & - \\ 
			SCSP~\cite{paper83}  & 24.2 & 44.6 & 54.1 & 65.2 & -\\  
			NLML~\cite{paper85} & 24.5 & 35.9 & 43.5 & 55.2 & - \\ 	
			GoG + XQDA~\cite{paper81}  & 24.7 & \bf{47.0} & \bf{58.4} & 69.0 & -\\  \hline
			\bf{Kernel X-CRC} & \bf{26.6} & 45.4 & 57.2 & \bf{69.7} & \bf{76.1}\\ \hline 
		\end{tabular}
		}
		\caption{Top ranked approaches on the GRID dataset.}
		\label{table:grid}		
\end{table}
\section{Conclusions}
In this work, we tackled the person re-identification problem using the proposed~\emph{Kernel Cross-View Collaborative Representation based Classification  (Kernel X-CRC)} approach.~\emph{Kernel X-CRC} represents probe and gallery images using collaborative representation coefficients that are robust to small-sample-size and balance the intra-camera representativeness with the inter-camera similarity. We performed an extensive experimental evaluation showing that the~\emph{Kernel X-CRC} successfully combines nonlinear modeling and multi-task learning. We also observed that working in a discriminative and low-dimensional subspace, the proposed method reaches outperforming results, obtaining the best~\emph{rank-1} matching rates for the two smaller datasets evaluated (PRID450S and GRID) and remaining within the two best approaches on the VIPeR and CUHK01 datasets.    
{\small
\bibliographystyle{ieee}
\bibliography{paper}

\begin{thebibliography}{10}\itemsep=-1pt

\bibitem{paper05}
L.~An, M.~Kafai, S.~Yang, and B.~Bhanu.
\newblock Reference-based person re-identification.
\newblock In {\em Advanced Video and Signal Based Surveillance (AVSS), 2013
  10th IEEE International Conference on}, pages 244--249, Aug 2013.

\bibitem{paper04}
L.~An, S.~Yang, and B.~Bhanu.
\newblock Person re-identification by robust canonical correlation analysis.
\newblock {\em Signal Processing Letters, IEEE}, 22(8):1103--1107, Aug 2015.

\bibitem{paper38}
B.-G. Apurva and S.~K. Shah.
\newblock A survey of approaches and trends in person re-identification.
\newblock {\em Image and Vision Computing}, 32(4):270 -- 286, 2014.

\bibitem{paper60}
A.~Argyriou, T.~Evgeniou, and M.~Pontil.
\newblock Convex multi-task feature learning.
\newblock {\em Machine Learning}, 73(3):243--272, 2008.

\bibitem{paper43}
Y.~Cai and M.~Pietikäinen.
\newblock Person re-identification based on global color context.
\newblock In {\em Computer Vision – ACCV 2010 Workshops}, volume 6468 of {\em
  Lecture Notes in Computer Science}, pages 205--215. Springer Berlin
  Heidelberg, 2011.

\bibitem{paper61}
R.~Caruana.
\newblock Multitask learning.
\newblock {\em Machine learning}, 28(1):41--75, 1997.

\bibitem{paper83}
D.~Chen, Z.~Yuan, B.~Chen, and N.~Zheng.
\newblock Similarity learning with spatial constraints for person
  re-identification.
\newblock In {\em Proceedings of the IEEE Conference on Computer Vision and
  Pattern Recognition}, pages 1268--1277, 2016.

\bibitem{paper88}
S.-Z. Chen, C.-C. Guo, and J.-H. Lai.
\newblock Deep ranking for person re-identification via joint representation
  learning.
\newblock {\em IEEE Transactions on Image Processing}, 25(5):2353--2367, 2016.

\bibitem{paper90}
Y.-C. Chen, W.-S. Zheng, and J.~Lai.
\newblock Mirror representation for modeling view-specific transform in person
  re-identification.
\newblock In {\em International Conference on Artificial Intelligence}, pages
  3402--3408. AAAI Press, 2015.

\bibitem{paper89}
D.~Cheng, Y.~Gong, S.~Zhou, J.~Wang, and N.~Zheng.
\newblock Person re-identification by multi-channel parts-based cnn with
  improved triplet loss function.
\newblock In {\em Proceedings of the IEEE Conference on Computer Vision and
  Pattern Recognition}, pages 1335--1344, 2016.

\bibitem{paper42}
D.~S. Cheng, M.~Cristani, M.~Stoppa, L.~Bazzani, and V.~Murino.
\newblock Custom pictorial structures for re-identification.
\newblock In {\em British Machine Vision Conference (BMVC)}, 2011.

\bibitem{paper11}
R.~F. de~Carvalho~Prates and W.~R. Schwartz.
\newblock Appearance-based person re-identification by intra-camera
  discriminative models and rank aggregation.
\newblock In {\em International Conference on Biometrics, {ICB} 2015,
  Phuket,Thailand, 19-22 May, 2015}, pages 65--72, 2015.

\bibitem{paper10}
R.~F. de~Carvalho~Prates and W.~R. Schwartz.
\newblock Appearance-based person re-identification by intra-camera
  discriminative models and rank aggregation.
\newblock In {\em International Conference on Biometrics, {ICB} 2015, Phuket,
  Thailand, 19-22 May, 2015}, pages 65--72, 2015.

\bibitem{paper48}
M.~Farenzena, L.~Bazzani, A.~Perina, V.~Murino, and M.~Cristani.
\newblock Person re-identification by symmetry-driven accumulation of local
  features.
\newblock In {\em IEEE Conference on Computer Vision and Pattern Recognition
  (CVPR)}, pages 2360 --2367, June 2010.

\bibitem{paper68}
D.~Gray, S.~Brennan, and H.~Tao.
\newblock Evaluating appearance models for recognition, reacquisition, and
  tracking.
\newblock In {\em 10th IEEE International Workshop on Performance Evaluation of
  Tracking and Surveillance (PETS)}, 2007.

\bibitem{paper55}
M.~T. Harandi, C.~Sanderson, R.~Hartley, and B.~C. Lovell.
\newblock Sparse coding and dictionary learning for symmetric positive definite
  matrices: A kernel approach.
\newblock In {\em Computer Vision--ECCV 2012}, pages 216--229. Springer, 2012.

\bibitem{paper93}
M.~Hirzer, C.~Beleznai, P.~M. Roth, and H.~Bischof.
\newblock Person re-identification by descriptive and discriminative
  classification.
\newblock In {\em Scandinavian conference on Image analysis}, pages 91--102.
  Springer, 2011.

\bibitem{paper87}
M.~Hirzer, P.~M. Roth, and H.~Bischof.
\newblock Person re-identification by efficient impostor-based metric learning.
\newblock In {\em Advanced Video and Signal-Based Surveillance (AVSS), 2012
  IEEE Ninth International Conference on}, pages 203--208. IEEE, 2012.

\bibitem{paper86}
M.~Hirzer, P.~M. Roth, M.~K{\"o}stinger, and H.~Bischof.
\newblock Relaxed pairwise learned metric for person re-identification.
\newblock In {\em European Conference on Computer Vision}, pages 780--793.
  Springer, 2012.

\bibitem{paper85}
S.~Huang, J.~Lu, J.~Zhou, and A.~K. Jain.
\newblock Nonlinear local metric learning for person re-identification.
\newblock {\em arXiv preprint arXiv:1511.05169}, 2015.

\bibitem{paper84}
C.~Jose and F.~Fleuret.
\newblock Scalable metric learning via weighted approximate rank component
  analysis.
\newblock {\em arXiv preprint arXiv:1603.00370}, 2016.

\bibitem{paper47}
S.~Karanam, Y.~Li, and R.~Radke.
\newblock Sparse re-id: Block sparsity for person re-identification.
\newblock In {\em Proceedings of the IEEE Conference on Computer Vision and
  Pattern Recognition Workshops}, pages 33--40, 2015.

\bibitem{paper92}
F.~M. Khan and F.~Br{\'{e}}mond.
\newblock Person re-identification for real-world surveillance systems.
\newblock {\em arXiv preprint arXiv:1607.05975}, 2016.

\bibitem{paper56}
M.~I. Khedher and M.~A. El~Yacoubi.
\newblock Local sparse representation based interest point matching for person
  re-identification.
\newblock In {\em Neural Information Processing}, pages 241--250. Springer,
  2015.

\bibitem{paper57}
M.~I. Khedher, M.~A. El~Yacoubi, and B.~Dorizzi.
\newblock Multi-shot surf-based person re-identification via sparse
  representation.
\newblock In {\em Advanced Video and Signal Based Surveillance (AVSS), 2013
  10th IEEE International Conference on}, pages 159--164. IEEE, 2013.

\bibitem{paper58}
E.~Kodirov, T.~Xiang, and S.~Gong.
\newblock Dictionary learning with iterative laplacian regularisation for
  unsupervised person re-identification.
\newblock In {\em Proceedings of the British Machine Vision Conference (BMVC)},
  pages 44.1--44.12. BMVA Press, September 2015.

\bibitem{paper71}
M.~Koestinger, M.~Hirzer, P.~Wohlhart, P.~M. Roth, and H.~Bischof.
\newblock Large scale metric learning from equivalence constraints.
\newblock 2012.

\bibitem{paper62}
J.~Li, H.~Zhang, L.~Zhang, X.~Huang, and L.~Zhang.
\newblock Joint collaborative representation with multitask learning for
  hyperspectral image classification.
\newblock {\em Geoscience and Remote Sensing, IEEE Transactions on},
  52(9):5923--5936, 2014.

\bibitem{paper67}
W.~Li and X.~Wang.
\newblock Locally aligned feature transforms across views.
\newblock In {\em Proceedings of the IEEE Conference on Computer Vision and
  Pattern Recognition}, pages 3594--3601, 2013.

\bibitem{paper45}
S.~Liao, Y.~Hu, X.~Zhu, and S.~Z. Li.
\newblock Person re-identification by local maximal occurrence representation
  and metric learning.
\newblock In {\em Proceedings of the IEEE Conference on Computer Vision and
  Pattern Recognition}, pages 2197--2206, 2015.

\bibitem{paper17}
G.~Lisanti, I.~Masi, A.~Bagdanov, and A.~Del~Bimbo.
\newblock Person re-identification by iterative re-weighted sparse ranking.
\newblock {\em Pattern Analysis and Machine Intelligence, IEEE Transactions
  on}, 37(8):1629--1642, Aug 2015.

\bibitem{paper08}
G.~Lisanti, I.~Masi, and A.~Del~Bimbo.
\newblock Matching people across camera views using kernel canonical
  correlation analysis.
\newblock In {\em Proceedings of the International Conference on Distributed
  Smart Cameras}, ICDSC '14, pages 10:1--10:6, New York, NY, USA, 2014. ACM.

\bibitem{paper18}
C.~Liu, S.~Gong, and C.~C. Loy.
\newblock On-the-fly feature importance mining for person re-identification.
\newblock {\em Pattern Recognition}, 47(4):1602 -- 1615, 2014.

\bibitem{paper70}
C.~Liu, S.~Gong, C.~C. Loy, and X.~Lin.
\newblock Person re-identification: What features are important?
\newblock In {\em Computer Vision--ECCV 2012. Workshops and Demonstrations},
  pages 391--401. Springer, 2012.

\bibitem{paper65}
A.~Ma, P.~Yuen, and J.~Li.
\newblock Domain transfer support vector ranking for person re-identification
  without target camera label information.
\newblock In {\em Proceedings of the IEEE International Conference on Computer
  Vision}, pages 3567--3574, 2013.

\bibitem{paper41}
B.~Ma, Y.~Su, and F.~Jurie.
\newblock Bicov: a novel image representation for person re-identification and
  face verification.
\newblock In {\em Proceedings of the British Machine Vision Conference}, pages
  57.1--57.11. BMVA Press, 2012.

\bibitem{paper40}
B.~Ma, Y.~Su, and F.~Jurie.
\newblock Local descriptors encoded by fisher vectors for person
  re-identification.
\newblock In A.~Fusiello, V.~Murino, and R.~Cucchiara, editors, {\em Computer
  Vision – ECCV 2012. Workshops and Demonstrations}, volume 7583 of {\em
  Lecture Notes in Computer Science}, pages 413--422. Springer Berlin
  Heidelberg, 2012.

\bibitem{paper66}
L.~Ma, X.~Yang, and D.~Tao.
\newblock Person re-identification over camera networks using multi-task
  distance metric learning.
\newblock {\em Image Processing, IEEE Transactions on}, 23(8):3656--3670, 2014.

\bibitem{paper81}
T.~Matsukawa, T.~Okabe, E.~Suzuki, and Y.~Sato.
\newblock Hierarchical gaussian descriptor for person re-identification.
\newblock In {\em Proceedings of the IEEE Conference on Computer Vision and
  Pattern Recognition}, pages 1363--1372, 2016.

\bibitem{paper14}
S.~Paisitkriangkrai, C.~Shen, and A.~van~den Hengel.
\newblock Learning to rank in person re-identification with metric ensembles.
\newblock In {\em Proceedings of the IEEE Conference on Computer Vision and
  Pattern Recognition}, pages 1846--1855, 2015.

\bibitem{paper80}
R.~Prates, M.~Oliveira, and W.~R. Schwartz.
\newblock Kernel partial least squares for person re-identification.
\newblock In {\em IEEE International Conference on Advanced Video and
  Signal-Based Surveillance (AVSS)}, 2016.

\bibitem{paper79}
R.~Prates and W.~R. Schwartz.
\newblock Kernel hierarchical pca for person re-identification.
\newblock In {\em 23th International Conference on Pattern Recognition, ICPR
  2016, Cancun, MEXICO, December 4-8, 2016.}, 2016.

\bibitem{paper74}
P.~M. Roth, M.~Hirzer, M.~Köstinger, C.~Beleznai, and H.~Bischof.
\newblock Mahalanobis distance learning for person re-identification.
\newblock In {\em Person Re-Identification}, Springer ACVPR, pages 247--267.
  2014.

\bibitem{paper07}
W.~Schwartz and L.~Davis.
\newblock Learning discriminative appearance-based models using partial least
  squares.
\newblock In {\em Computer Graphics and Image Processing, 2009 XXII Brazilian
  Symposium on}, pages 322--329, Oct 2009.

\bibitem{paper64}
S.~Shekhar, V.~M. Patel, N.~M. Nasrabadi, and R.~Chellappa.
\newblock Joint sparse representation for robust multimodal biometrics
  recognition.
\newblock {\em Pattern Analysis and Machine Intelligence, IEEE Transactions
  on}, 36(1):113--126, 2014.

\bibitem{paper50}
C.~Su, F.~Yang, S.~Zhang, Q.~Tian, L.~S. Davis, and W.~Gao.
\newblock Multi-task learning with low rank attribute embedding for person
  re-identification.
\newblock In {\em International Conference on Computer Vision}, pages
  3739--3747, 2015.

\bibitem{paper75}
C.~Tian, M.~Zeng, and Z.~Wu.
\newblock Person re-identification based on spatiogram descriptor and
  collaborative representation.
\newblock {\em IEEE Signal Processing Letters}, 22(10):1595--1599, 2015.

\bibitem{paper53}
J.~Wright, Y.~Ma, J.~Mairal, G.~Sapiro, T.~S. Huang, and S.~Yan.
\newblock Sparse representation for computer vision and pattern recognition.
\newblock {\em Proceedings of the IEEE}, 98(6):1031--1044, 2010.

\bibitem{paper72}
S.~Wu, Y.-C. Chen, X.~Li, J.-J. You, and W.-S. Zheng.
\newblock An enhanced deep feature representation for person re-identification.
\newblock In {\em WACV2016: IEEE Winter Conference on Applications of Computer
  Vision.}, March 2016.

\bibitem{paper59}
M.~Yang, L.~Zhang, D.~Zhang, and S.~Wang.
\newblock Relaxed collaborative representation for pattern classification.
\newblock In {\em Computer Vision and Pattern Recognition (CVPR), 2012 IEEE
  Conference on}, pages 2224--2231. IEEE, 2012.

\bibitem{paper44}
Y.~Yang, J.~Yang, J.~Yan, S.~Liao, D.~Yi, and S.~Z. Li.
\newblock Salient color names for person re-identification.
\newblock In {\em Computer Vision--ECCV 2014}, pages 536--551, 2014.

\bibitem{paper54}
X.-T. Yuan, X.~Liu, and S.~Yan.
\newblock Visual classification with multitask joint sparse representation.
\newblock {\em Image Processing, IEEE Transactions on}, 21(10):4349--4360,
  2012.

\bibitem{paper76}
M.~Zeng, Z.~Wu, C.~Tian, L.~Zhang, and L.~Hu.
\newblock Efficient person re-identification by hybrid spatiogram and
  covariance descriptor.
\newblock In {\em IEEE Conference on Computer Vision and Pattern Recognition
  Workshops}, pages 48--56, 2015.

\bibitem{paper63}
H.~Zhang, N.~M. Nasrabadi, Y.~Zhang, and T.~S. Huang.
\newblock Joint dynamic sparse representation for multi-view face recognition.
\newblock {\em Pattern Recognition}, 45(4):1290--1298, 2012.

\bibitem{paper82}
L.~Zhang, T.~Xiang, and S.~Gong.
\newblock Learning a discriminative null space for person re-identification.
\newblock {\em arXiv preprint arXiv:1603.02139}, 2016.

\bibitem{paper52}
L.~Zhang, M.~Yang, X.~Feng, Y.~Ma, and D.~Zhang.
\newblock Collaborative representation based classification for face
  recognition.
\newblock {\em arXiv preprint arXiv:1204.2358}, 2012.

\bibitem{paper91}
Y.~Zhang, B.~Li, H.~Lu, A.~Irie, and X.~Ruan.
\newblock Sample-specific svm learning for person re-identification.
\newblock In {\em The IEEE Conference on Computer Vision and Pattern
  Recognition (CVPR)}, June 2016.

\bibitem{paper46}
R.~Zhao, W.~Ouyang, and X.~Wang.
\newblock Unsupervised salience learning for person re-identification.
\newblock In {\em Computer Vision and Pattern Recognition (CVPR), 2013 IEEE
  Conference on}, pages 3586--3593, June 2013.

\bibitem{paper73}
W.-S. Zheng, S.~Gong, and T.~Xiang.
\newblock Person re-identification by probabilistic relative distance
  comparison.
\newblock In {\em Computer vision and pattern recognition (CVPR), 2011 IEEE
  conference on}, pages 649--656. IEEE, 2011.

\end{thebibliography}
}

\end{document}